\pdfoutput=1

\documentclass[11pt]{article}

\usepackage[]{EMNLP2023}

\usepackage{times}
\usepackage{latexsym}
\usepackage{amsmath}
\usepackage[T1]{fontenc}

\usepackage[utf8]{inputenc}

\usepackage{microtype}

\usepackage{inconsolata}

\usepackage{graphicx}
\usepackage{epstopdf}
\usepackage{diagbox}
\usepackage{booktabs}
\usepackage{enumerate}

\title{Summarization is (Almost) Dead}

\author{Xiao Pu$^*$, Mingqi Gao$^*$, Xiaojun Wan \\
Wangxuan Institute of Computer Technology, Peking University \\
\texttt{puxiao@stu.pku.edu.cn}\\
\texttt{\{gaomingqi, wanxiaojun\}@pku.edu.cn}\\}

\begin{document}
\maketitle
\def\thefootnote{*}\footnotetext{Equal contribution.}\def\thefootnote{\arabic{footnote}}

\begin{abstract}
How well can large language models (LLMs) generate summaries? We develop new datasets and conduct human evaluation experiments to evaluate the zero-shot generation capability of LLMs across five distinct summarization tasks. Our findings indicate a clear preference among human evaluators for LLM-generated summaries over human-written summaries and summaries generated by fine-tuned models. Specifically, LLM-generated summaries exhibit better factual consistency and fewer instances of extrinsic hallucinations.
Due to the satisfactory performance of LLMs in summarization tasks (even surpassing the benchmark of reference summaries), we believe that most conventional works in the field of text summarization are no longer necessary in the era of LLMs. However, we recognize that there are still some directions worth exploring, such as the creation of novel datasets with higher quality and more reliable evaluation methods.




\end{abstract}

\section{Introduction}

Text summarization, a natural language generation (NLG) task, aims to compress extensive source materials into brief summaries, including diverse content such as news articles, source codes, and cross-lingual text. Traditional methods used fine-tuning techniques on specific datasets \citep{lewis2019bart,raffel2020exploring,zhang2020pegasus,wang2021codet5,xue2021mt5,wang2022clidsum}, but the emergence of large language models (LLMs) \citep{chung2022scaling,zhang2022opt,touvron2023llama,ouyang2022training,openai2023gpt4} has shifted the focus to their promising zero-shot generation capability.

In an exploration of this potential, we evaluated the performance of LLMs on various summarization tasks—single-news, multi-news, dialogue, source codes, and cross-lingual summarization—using human-generated evaluation datasets. Our quantitative and qualitative comparisons between LLM-generated summaries, human-written summaries, and summaries generated by fine-tuned models revealed that \textbf{LLM summaries are significantly preferred by the human evaluators, which also demonstrate higher factuality}.

Given the impressive performance of LLMs across these tasks, we question the need for further refinement of text summarization models with higher metric scores. After sampling and examining 100 summarization-related papers published in ACL, EMNLP, NAACL, and COLING in the past 3 years, we find that the main contribution of about 70\% papers was to propose a summarization approach and validate its effectiveness on standard datasets. As such, we provocatively assert that " Summarization is (almost) Dead." Nonetheless, we acknowledge existing challenges in the field such as the need for high-quality reference datasets, application-oriented approaches, and improved evaluation methods.

\begin{figure*}[t]
    \centering
    \includegraphics[width=\linewidth]{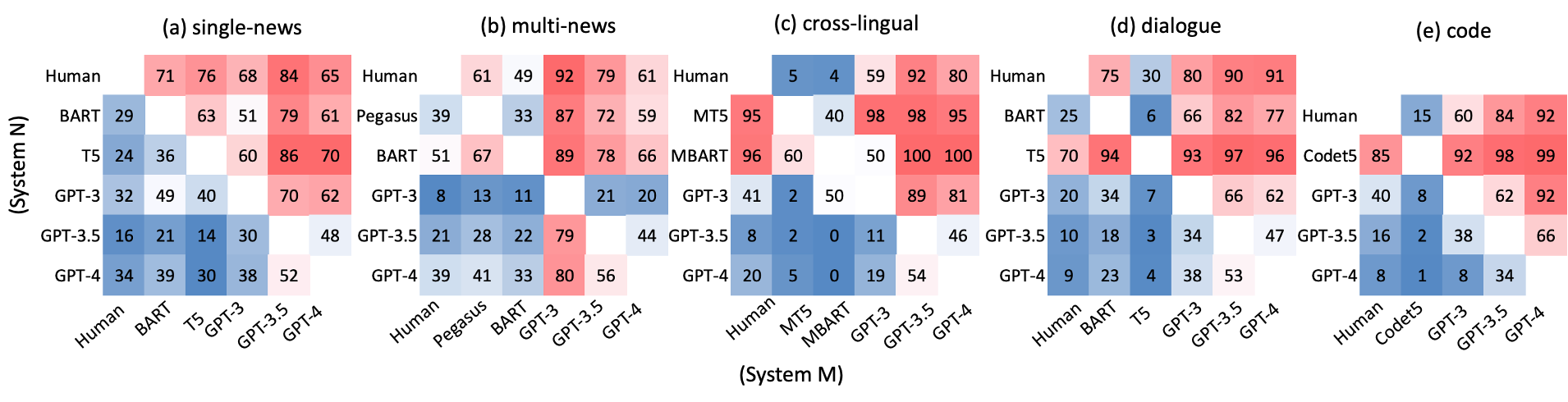}
    \vspace{-5mm}
    \caption{Pairwise winning rates (\%) between different systems across 5 tasks. Each data point represents the proportion of times System M (horizontal axis) is preferred over System N (vertical axis) in the comparisons. Red indicates a winning rate greater than 50\%, indicating a preference for System M, while blue represents a winning rate less than 50\%, indicating a preference for System N, with darker colors indicating the greater difference in winning rates between the two systems.We find that LLMs are highly preferred by human evaluators.}
    \label{fig:winning_rate}
\end{figure*}


\section{Experimental Settings}
In this section, we provide an overview of the datasets and models used for human evaluation, as well as the experimental process and  details.
\subsection{Datasets}
To ensure that the large language model has not "seen" the data during training, we use the latest data to build the datasets specifically for human evaluation in each summarization task.\footnote{According to OpenAI's disclosure, the training data for GPT-3, GPT-3.5, and GPT-4 are cut off until September 2021. Therefore, our datasets are sourced from a time point after that cutoff date.} Each dataset consists of 50 samples.

In conducting single-news, multi-news, and dialogue summarization tasks, our methodology emulates the dataset construction approaches utilized by CNN/DailyMail \citep{see-etal-2017-get, hermann2015teaching}, Multi-News \citep{fabbri-etal-2019-multi}, and Mediasum \citep{zhu2021mediasum}, respectively, to build the datasets for our experiments. The sources for the data remain consistent with the original datasets, such as the DailyMail website, yet are the latest. For the cross-lingual summarization task, our strategy aligns with \citet{zhu2019ncls} approach, which entails translating reference summaries in our single-news dataset from English to Chinese using Google Translate, followed by a post-editing process. \footnote{Post-editing is done by a graduate student in linguistics.} In relation to code summarization, we adopt \citet{bahrami2021pytorrent}'s methodology to formulate a dataset, the source documents of which are Go language programs.

\subsection{Models}
For each task, we choose GPT-3(text-davinci-003) \citep{brown2020language,ouyang2022training}, GPT-3.5 and GPT-4 \citep{openai2023gpt4} as representatives of LLMs. For each summarization task, we additionally utilize 1-2 smaller models, previously fine-tuned on a dataset in this specific task. Precisely, we employ BART \citep{lewis2019bart} and T5 \citep{raffel2020exploring} for single-news task, Pegasus \citep{zhang2020pegasus} and BART for multi-news task, T5 and BART for Dialogue task, MT5 \citep{xue2021mt5} and MBART \citep{wang2022clidsum} for cross-lingual task and Codet5 \citep{wang2021codet5} for source code task.

\subsection{Experimental process and details}
We conduct human evaluation experiments to annotate the different types of summaries for each of five tasks. For each task, we hire two annotators \footnote{
We recruit graduate students with a good command of English to participate in our experiments. For the code summarization task, we hire two students majoring in computer science who are familiar with the Go programming language.}, and each annotator is assigned to complete all 50 questions of a single task. For each question, they are presented with a source article and summaries from all summarization systems selected in this task. They are then asked to compare the summaries pairwise. If there are a total of $n$ systems in the task, each annotator will need to make $C_{n}^{2}$ comparisons for one question. We calculate the overall 
Cohen's kappa coefficient \citep{cohen1968weighted} and find that the inter-annotator agreement is acceptable, with a coefficient of 0.558.

\section{Experiment Results}
\subsection{Experiment 1: Comparing the overall quality of summaries}
In this experiment, we engage human evaluators to compare the overall quality of different summaries. Then we compute $\text{Win\-Rate}_{M}^{N}$, which represents the proportion of times system M is preferred by human evaluators when comparing it with system N. 
By comparing the pairwise winning rates among different systems, we can gain insights into the relative overall quality of the systems. Surprisingly, as depicted in Figure 1, \textbf{summaries generated by the LLMs consistently outperform both human and summaries generated by fine-tuned models across all tasks.}

This raises the question of why LLMs are able to outperform human-written summaries, which are traditionally regarded as flawless and oracle. Moreover, it prompts us to examine the specific limitations of human-written references. Our initial observations suggest that LLM-generated summaries exhibit high fluency and coherence. However, the comparative factual consistency between LLM summaries and human-written ones remains uncertain. Consequently, our next experiment focuses on exploring the factual consistency aspect.

\subsection{Experiment 2: Comparing the factual consistency of summaries}
We further recruit annotators to identify sentence-level hallucinations in the human- and LLM-generated summaries, allowing us to compare their levels of factual consistency. Given the significant cost of annotation, we select GPT-4 as a representative LLM.
As depicted in Table \ref{tab:factual}, human-written reference summaries exhibit either an equal or higher number of hallunications compared to GPT-4 summaries. In specific tasks such as multi-news and code summarization, human-written summaries exhibit notably inferior factual consistency. 

\begin{table}[t]
\resizebox{\columnwidth}{!}{%
\begin{tabular}{l|llllll}
\toprule
\textbf{System} & \begin{tabular}[c]{@{}l@{}}Single-\\ news\end{tabular} & \begin{tabular}[c]{@{}l@{}}Multi-\\ news\end{tabular} & \begin{tabular}[c]{@{}l@{}}Cross-\\ lingual\end{tabular} & Dialogue & Code \\
\midrule
\textbf{GPT-4}& 8 & 5 & 16 & 5& 9 \\
\textbf{Human}& 13 & \textbf{62} & 15 & 15 & \textbf{46} \\
\bottomrule
\end{tabular}%
}
\caption{The number of hallucinations (sentence-level) found in GPT-4 and human-written summaries. We highlight the figures which is significantly large.
}
\label{tab:factual}
\end{table}

To gain a deeper understanding of this observed phenomenon, we further investigate the types of these factual errors. Following \citet{maynez-etal-2020-faithfulness} we divide all hallucinations into two categories: intrinsic and extrinsic hallucinations. Intrinsic hallucinations refer to inconsistencies between the factual 
information in the summary and the source text, while extrinsic hallucinations occur when the summary includes certain factual information that is not present in the source text.

By analyzing the proportion of intrinsic and extrinsic hallucinations in both human-written and GPT-4 summaries, as shown in Table \ref{tab:ex-halluc}, we discover \textbf{a notably higher occurrence of extrinsic hallucinations in tasks where human-written summaries demonstrate poor factual consistency}, e.g., multi-news and code. Howerver, where there is little difference in factual consistency between human and GPT-4, the proportions of extrinsic hallucinations in both systems are similar. Therefore, we hypothesize that extrinsic hallucinations are the underlying cause for the inadequate factual consistency observed in human-written summaries.
\begin{table}[t]
\resizebox{\columnwidth}{!}{%
\begin{tabular}{l|l|lllll}
\toprule
\textbf{System} & Avg & \begin{tabular}[c]{@{}l@{}}Single-\\ news\end{tabular} & \begin{tabular}[c]{@{}l@{}}Multi-\\ news\end{tabular} & \begin{tabular}[c]{@{}l@{}}Cross-\\ lingual\end{tabular} & Dialogue & Code \\
\midrule
\textbf{GPT-4} & 40\% & 50\% & \textbf{40\%} & 38\% & 40\% & \textbf{33\%} \\
\textbf{Human} & 62\% & 62\% & \textbf{73\%} & 33\% & 53\% & \textbf{89\%} \\
\bottomrule
\end{tabular}%
}
\caption{The proportion of extrinsic hallucinations in GPT-4 and human-written summaries. 
}
\label{tab:ex-halluc}
\end{table}

\subsection{Comparative Analysis}
Here we delve into an analysis of the specific strengths exhibited by LLM summaries in comparison to both human and fintuned summaries. We have included some concrete examples and a more detailed version of analysis in Appendix \ref{appsec:case_study}.
\subparagraph{Reference summaries vs. LLM summaries}
Compared to LLM summaries, we identify a specific issue with human-written reference summaries, their lack of fluency. As shown in Figure \ref{fig:case_study_human}(a), human-written reference summaries are sometimes flawed with incomplete information. Another issue observed in some human-written reference summaries, as concluded in the previous chapter's quantitative analysis and shown in Figure \ref{fig:case_study_human}(b) , is the presence of hallucinations.
\subparagraph{Summaries generated by fine-tuned models vs. LLM summaries}
In comparison to LLM summaries, we find that summaries generated by fine-tuned models tend to have a fixed and rigid length, whereas LLMs are able to adjust the output length according to the input's information volume. Additionally, when the input contains multiple topics, the summaries generated by fine-tuned models demonstrate lower coverage of these topics, as exampled in Figure \ref{fig:case_study_ft}, while LLMs can capture all the topics when generating summaries.

\section{The Changing Landscape of Summarization: Seeking New Horizons}


Through the aforementioned manual evaluation, we have discovered that the quality of summaries generated by LLMs surpasses that of the reference summaries in many datasets. It is foreseeable that with the continuous improvement of future LLMs, their summarization capabilities will further enhance. Previous summarization methods were often tailored to specific categories, domains, or languages, resulting in limited generality, and their significance is gradually diminishing. As mentioned in the introduction, nearly 70\% of the research is no longer meaningful. However, we believe that the following directions are worth exploring:

\subsection{Summarization Datasets}

The role of the dataset shifts from model training to testing, necessitating higher-quality reference summaries. Previously generated datasets will gradually be phased out, and future reference summaries will require human expert annotations. 

The majority of current summarization datasets are in English and focused on news articles, scientific articles, or Wikipedia \citep{wikidump,merity2016pointer}. And the source documents are relatively short. In order to thoroughly assess the summarization capabilities of LLMs, it becomes imperative to incorporate other \textbf{diverse genres} of data, as well as \textbf{other languages}, especially those that are low-resource in nature. Additionally, there is a need to include \textbf{longer documents}, such as books, within the datasets to facilitate comprehensive evaluation.




\subsection{Summarization Approaches}

Further investigation is warranted in the realm of application-oriented summarization approaches, with the assistance of LLMs.

\textbf{Customized Summarization} \citep{zhong-etal-2022-unsupervised}: LLMs can be customized to generate summaries that align with individual user preferences, reading history, or expertise level, thereby personalizing the summarization process.

\textbf{Real-time Summarization} \citep{yang2022catchlive}: The ability to condense information in real time plays a vital role in various contexts, such as live steams, stock market fluctuations, or social media monitoring. Research efforts in this domain could concentrate on enhancing the promptness and efficiency of LLMs.

\textbf{Interactive Summarization} \cite{shapira-etal-2017-interactive}: The development of models capable of interacting with users, soliciting clarification or feedback throughout the summarization process, holds promise for augmenting the accuracy and relevance of summaries.

\subsection{Summarization Evaluation}
It is imperative to bid farewell to antiquated assessment metrics such as ROUGE \citep{lin-2004-rouge}, as they no longer align with the evolving landscape of summarization. Future automated evaluation techniques for summarization hold promise in their reliance on LLMs as demonstrated by recent studies \citep{kocmi2023large, wang2023chatgpt, liu2023gpteval, luo2023chatgpt, gao2023humanlike}. Furthermore, a shift in emphasis is necessary in the evaluation of summarization, with greater consideration given to the practical utility and applications of the generated summaries.

\textbf{Extrinsic Evaluation} \citep{pu2023summary}: Measuring the effectiveness of a summary by using it as an input to another task (e.g., question-answering or decision-making tasks), to see if essential information has been retained.

\section{Related Work}
\textbf{Evaluating the ability of LLMs on summarization}. \citet{goyal2023news} show that news summaries generated by GPT-3 are overwhelmingly preferred by humans compared with those generated by fine-tuned models. Further, \citet{zhang2023benchmarking} find that news summaries generated by LLMs are evaluated to be on par with human-written summaries. Some studies have also explored the performance of LLMs such as ChatGPT with automatic evaluations on aspect-based summarization \citep{yang2023exploring} and cross-lingual summarization \citep{wang2023elementaware}. Our work involves GPT-3.5 and GPT-4 and conducts human evaluations on various data on which they have not been trained.



\section{Conclusion}
Through the development of new evaluation datasets and the conduction of comprehensive human evluation experiments that cover a wide range of summarization scenarios, our study demonstrates the remarkable performance of LLM-generated summaries compared to human-written reference summaries and summaries generated by fine-tuned models across diverse summarization tasks. LLM summaries exhibit superior fluency, factuality, and flexibility, especially in specialized and uncommon summarization scenarios. Our findings indicate that text summarization is undergoing significant transformation, rendering previous approaches less meaningful in light of the advancements in LLMs. We also offer an outlook on the tasks worth exploring in the field of text summarization in the future, focusing on three aspects: datasets, methods, and evaluation. We hope that this can bring inspiration to relevant researchers regarding their future work.

\section*{Limitations}
We do not include other popular LLMs like LLaMA and Vacuna because these newer models do not disclose the cutoff date of their training data. This lack of information makes it challenging for us to create a novel dataset specifically tailored for evaluating the zero-shot generation of summaries by LLMs. In the future, if these LLMs provide more details about their training data, it would be beneficial to incorporate additional LLMs and compare the differences among different series of such models.

Due to the high cost, we only conduct human experiments on five common text summarization tasks. However, exploring some less common and more challenging summarization tasks, e.g. slides summarization, to test the capability of LLMs in the field text summarization would be an interesting avenue for future research. 

\section*{Ethics Statement}

All annotators are paid 10 USD per hour, above the local minimum wage. During the process of constructing the dataset, we obtain text from publicly available websites, and it is possible that some of the texts may contain biased, offensive or violent elements. However, it is important to note that the inclusion of such content does not reflect our research standpoint or endorse any biased, offensive or violent views. Our intention is to use the dataset only for research purposes.


\bibliography{emnlp2023}
\bibliographystyle{acl_natbib}

\appendix
\section{Case Studies}
\label{appsec:case_study}
\begin{figure*}[t]
    \centering
    \includegraphics[width=\linewidth]{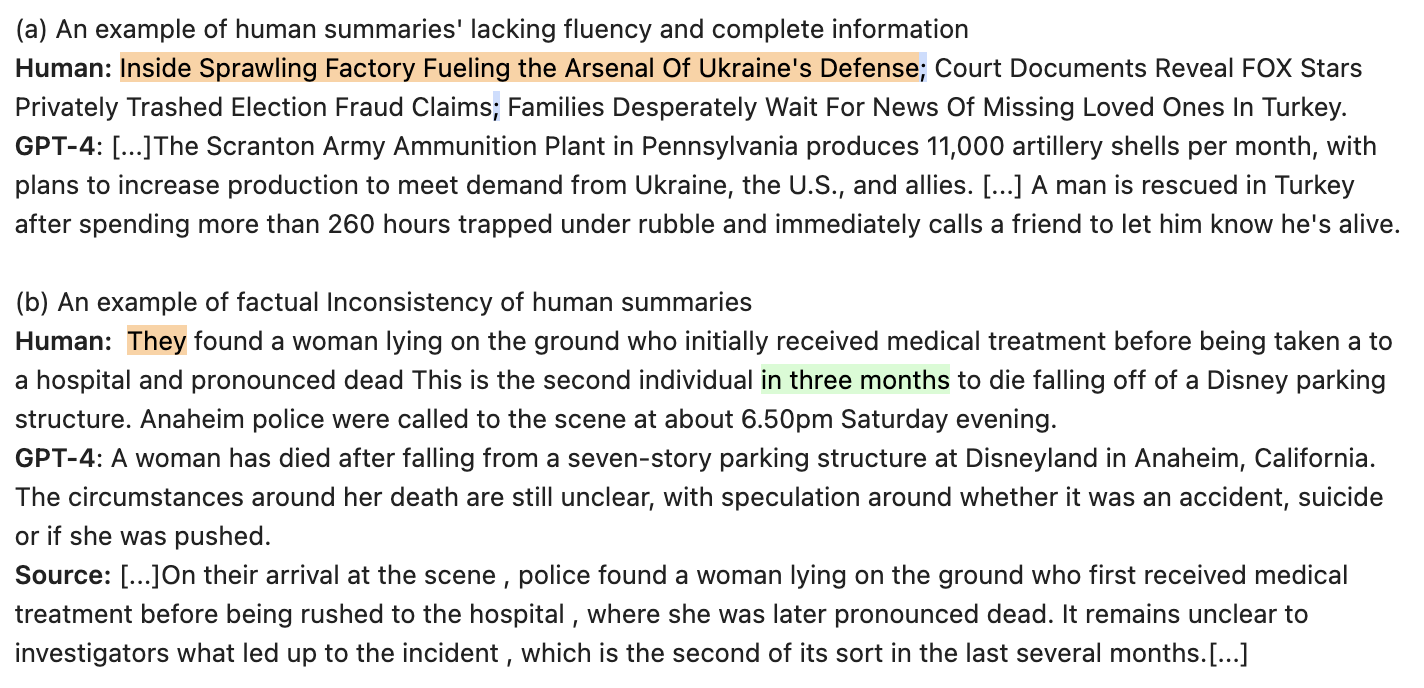}
    \caption{A case study comparing human-written reference and GPT-4 summaries, we have color-coded the issues in the human summaries for better clarity: orange for incomplete information, blue for broken sentences, and green for hallucinations.}
    \label{fig:case_study_human}
\end{figure*}

\textbf{Compared to LLM, the weaknesses of human-written reference summaries lie in incomplete information and hallucinations.} As shown in Figure \ref{fig:case_study_human}(a), human-written reference summaries are sometimes flawed with incomplete information, where we are only informed about the event without knowledge of the involved participants. In addition, some references lack fluency resulting in decreased readability. Another issue observed in some human-written reference summaries, as concluded in the previous chapter's quantitative analysis, is the presence of factual errors or hallucination. For instance, source text of Figure \ref{fig:case_study_human}(b) only mentioned two incidents occurring within several months, while the human-written reference summary incorrectly stated "three months." 

\begin{figure*}[tbh]
    \centering
    \includegraphics[width=\linewidth]{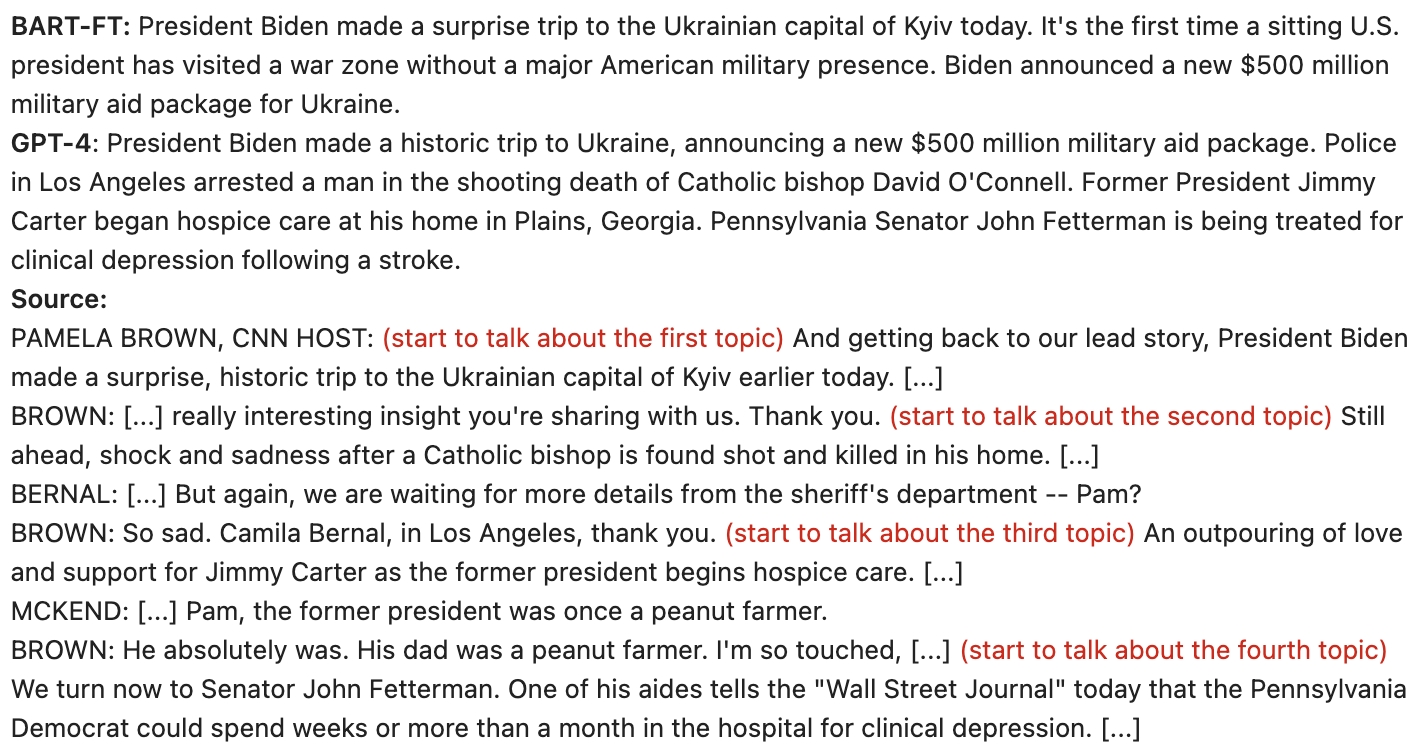}
    \caption{A case study comparing fine-tuned and GPT-4 summaries. The dialogue is a transcript of a news program where a total of four topics are discussed. For the source text, we annotate the transitions between dialogue topics in red.}
    \label{fig:case_study_ft}
\end{figure*}

\textbf{Compared to LLM summaries, the drawbacks of fine-tuned summaries lie in their inability to effectively capture all the topics mentioned in a long article.} They tend to focus on only a small sub-set of the topics. As shown in Figure \ref{fig:case_study_ft}, the source text discusses four news events sequentially, while the fine-tuned BART summaries only covers the information of the first news event. In contrast, GPT-4 successfully summarizes all four news events within a similar length.

\section{Human preference score}
\label{appsec:case_study}
\begin{figure*}[tbh]
    \centering
    \includegraphics[width=\linewidth]{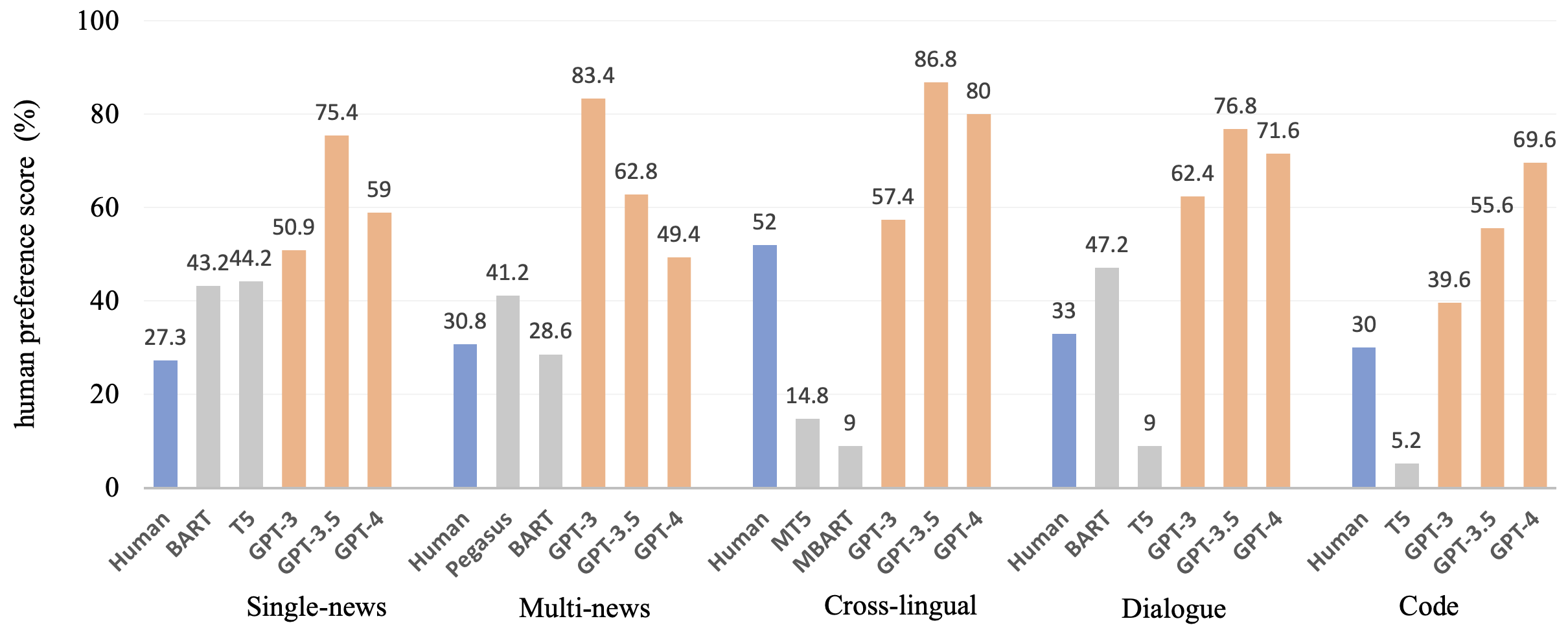}
    \caption{Human preference scores for different summary systems across 5 tasks. LLMs are highly preferred by human evaluators, as demonstrated by the higher scores assigned to LLM-generated summaries compared to other systems.}
    \label{fig:overall_scores}
\end{figure*}

Apart from pairwise comparisons of the performance of different systems, we also introduce human preference score to indicate the relative quality of a system in each task. It equals to the overall winning rate of the system in comparisons with any other systems, based on annotators' judgements. As shown in Figure \ref{fig:overall_scores}, the human preference scores for the large models exceed 50\%, indicating a strong preference for their summaries and highlighting the capability of LLMs in text summarization.
\end{document}